\documentclass[10pt,twocolumn,letterpaper]{article}

\usepackage{iccv}
\usepackage{times}
\usepackage{epsfig}
\usepackage{graphicx}
\usepackage{amsmath}
\usepackage{amssymb}
\usepackage{enumitem}

\def\R{\mathbf{R}}
\def\D{\mathbf{D}}

\def\B{\mathbf{B}}

\def\supi{{(i)}}

\def\bz{\mathbf{z}}
\def\bw{\mathbf{w}}
\def\M{\mathbf{M}}
\def\S{\mathbf{S}}
\def\W{\mathbf{W}}

\def\cL{\mathcal{L}}
\def\bvarphi{\boldsymbol{\varphi}}
\def\btheta{\boldsymbol{\theta}}

\usepackage[pagebackref=true,breaklinks=true,letterpaper=true,colorlinks,bookmarks=false]{hyperref}

\iccvfinalcopy % *** Uncomment this line for the final submission

\def\iccvPaperID{6608} % *** Enter the ICCV Paper ID here

\ificcvfinal\fi
\begin{document}

%%%%%%%%% TITLE
\title{Distill Knowledge from NRSfM for Weakly Supervised 3D Pose Learning}

\author{Chaoyang Wang~~~~Chen Kong~~~~Simon Lucey\\
Carnegie Mellon University\\
{\tt\small \{chaoyanw, chenk, slucey\}@cs.cmu.edu}
% \and
}

\maketitle
%\thispagestyle{empty}

%%%%%%%%% ABSTRACT
\begin{abstract}
% Weakly supervised 3D Pose Estimation has been explored under different settings, e.g. 2D landmark annotation with extra information and constraints. 

We propose to learn a 3D pose estimator by distilling knowledge from Non-Rigid Structure from Motion (NRSfM). Our method uses solely 2D landmark annotations. No 3D data, multi-view/temporal footage, or object specific prior is required. This alleviates the data bottleneck, which is one of the major concern for supervised methods. The challenge for using NRSfM as teacher is that they often make poor depth reconstruction when the 2D projections have strong ambiguity. Directly using those wrong depth as hard target would negatively impact the student. Instead, we propose a novel loss that ties depth prediction to the cost function  used in NRSfM. This gives the student pose estimator freedom to reduce depth error by associating with image features. Validated on H3.6M dataset, our learned 3D pose estimation network achieves more accurate reconstruction compared to NRSfM methods. It also outperforms other weakly supervised methods, in spite of using significantly less supervision.

% It also achieves estimation accuracy comparable to other weakly supervised methods, in spite of using significantly less supervision.

%In this paper, we propose an algorithm work with the minimum settings - we use only 2D annotation, no partial/unpaired 3D annotation, no multi-view/temporal information, and no object specific kinematic priors. 

%Our algorithm is derived from non-rigid structure from motion (NRSfM), but differs in that we enforce image information (translated by 3D pose estimation network) as a hard constraint onto the NRSfM optimization objective. Our image constrained NRSfM (IC-NRSfM) is able to better disambiguate poses which are difficult %for NRSfM using only 2D projections. To optimize this IC-NRSfM learning objective, we propose a novel optimization procedure which allows end-to-end training. In experiment, both our NRSfM baseline and IC-NRSfM outperforms existing NRSfM methods by a large margin on H3.6M dataset; and on H3.6M validation set, our pose estimation accuracy is close to state-of-the-arts weakly supervised methods using stricter settings.
% \cy{rewrite to match current narrative}
\end{abstract}

%%%%%%%%% BODY TEXT

\section{Introduction}

Learning to estimate 3D pose from images is bottlenecked by the availability of abundant 3D annotated data. Weakly supervised methods that reduce the amount of required annotation is of high practical value. Prior works approach this problem by supplementing their training set with: (i) extra 2D annotated data~\cite{sup3}; (ii) aligning 3D models to 2D annotations~\cite{song2018apollocar3d,w_3dinterp,Tome_2017_CVPR}; (iii) exploiting geometric cues from multi-view footage~\cite{w_Rhodin_cvpr,w_Rhodin_eccv,tung2017self}; or (iv) utilizing adversarial framework to impose a prior on the 3D structure~\cite{drover2018can}. These methods, however, are either restricted to laboratory settings or still requires a 3D training set -- which limits the type of target objects they can work with. This paper addresses a more general setting -- we utilize image datasets with solely 2D landmark annotations (i.e. no 3D supervision). This allows our method to be applied to a wider scope of objects, not limited by the availability of 3D models, kinematic priors, or sequential/multi-view footage. 

\begin{figure}[t]
    \centering
    \includegraphics[width=\linewidth]{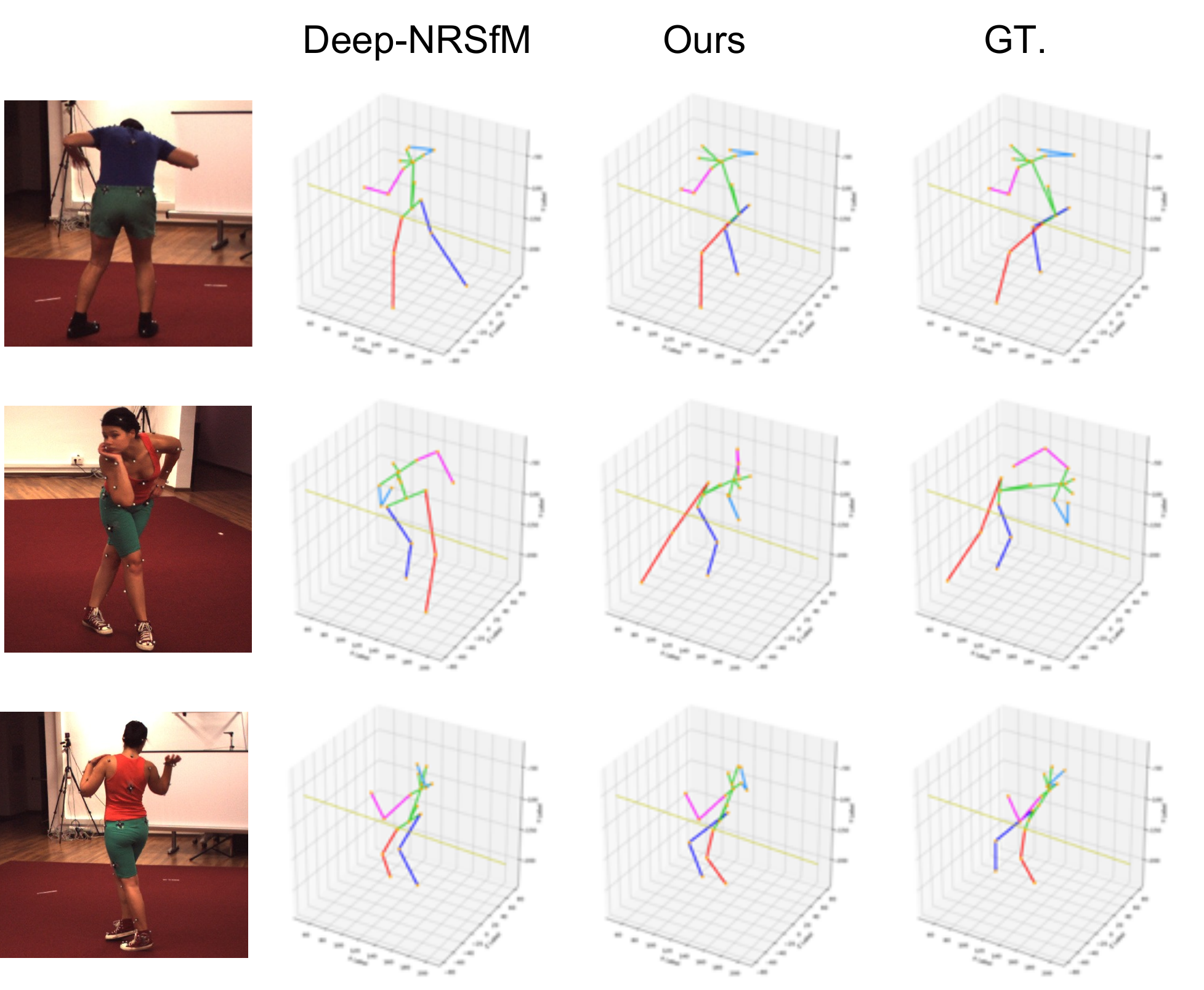}
    \caption{NRSfM methods often achieve poor reconstructions when the 2D projections have strong ambiguity. Our proposed knowledge distilling method lets the student pose estimation network (3rd column) correct some of the mistakes made by its NRSfM teacher (2nd column). }
    \label{fig:teaser}
\end{figure}

Our work is made possible by some recent advances in Non-Rigid Structure from Motion (NRSfM). NRSfM methods reconstruct 3D shapes and camera positions from multiple 2D projections of articulated 3D points. These points do not have to belong to the same object, but can be from multiple instances of the same object category, which naturally applies to our problem. Prior NRSfM methods are restricted by the number of frames and the type of shape variability they can handle, which limits their usage to many real world problems. Kong and Lucey~\cite{ck19} recently proposed a neural network architecture (Deep-NRSfM) interpreted as solving a multi-layer block sparse dictionary learning problem, and can handle problems of unprecedented scale and shape complexity. Our \textit{modified} version of Deep-NRSfM achieves state-of-the-arts accuracy on H3.6M~\cite{h36m} dataset, outperforming other NRSfM methods by a significant margin.

Despite this progress, NRSfM still has difficulty in predicting correct depth for shapes with strong ambiguity in terms of 2D projection, e.g. identifying if a leg is stretching towards/away from the camera, even though these are distinguishable with texture features. Therefore, directly using the depth output from NRSfM as labels to train a pose estimation network is affected by those errors. Instead of this hard assignment of training labels, we propose a softer approach -- we want to penalize less when there's high ambiguity in 2D projection, so as to leave room for the pose estimation network to correct errors made by NRSfM through associating image features (see Fig.~\ref{fig:teaser}).%As shown in Fig.~\ref{fig:teaser}, our approach allows the pose estimation network to even correct those errors made by NRSfM on the training set.

To design our learning objective, we review the dictionary learning problem used to solve NRSfM. Assuming the camera matrix fixed, a depth hypothesis defines a subspace of codes -- any codes in this subspace is to have the same depth reconstruction as the hypothesis, but have different cost (2D reprojection error + regularizer). A natural way to characterize the quality of a depth hypothesis is by the minimum cost of codes in its subspace. However, directly using this as a learning objective leads to solving a constrained optimization problem numerically per SGD iteration, which is computationally intractable. Instead, we derive a convex upper bound by evaluating the cost at the projection of the NRSfM solution on the subspace. Experiments show that pose network trained by this loss noticeably reduces error on the training set compared to our already strong NRSfM baseline, and consequently leads to lower validation error as a weakly supervised learning task. 

Another benefit of the proposed knowledge distilling loss is that, it poses no restriction on the architecture of the student pose estimation network, as long as it outputs the depth value for the landmarks. This is not the case for some of the prior works~\cite{w_3dinterp,w_aign}, where the pose estimation network has to output the coefficients associated to some external shape dictionary.

In conclusion, contributions of this paper are:
\begin{itemize}[topsep=0pt, partopsep=0pt, itemsep=0pt,parsep=2pt]
    \item We propose a weakly supervised pose estimation method using solely 2D landmark annotations. We do not use any 3D labels, multi-view footage, or target specific shape prior. In spite of using weaker supervision, we achieve the best results compared to other weakly supervised methods.
    \item We establish a strong NRSfM baseline modified from Deep-NRSfM~\cite{ck19}, which outperforms current published state-of-the-art NRSfM methods on H3.6M dataset.
    \item We propose a new knowledge distilling algorithm applicable to NRSfM methods based on dictionary learning. We demonstrate that our learned network gets significantly lower error on the training set compared to its NRSfM teacher. 
\end{itemize}

\section{Related Works}

\paragraph{Non-rigid structure from motion}
NRSfM is a classical ill-posed problem since the 3D shapes can vary between
images, resulting in more variables than equations. To alleviate the
ill-posedness, various constraints are exploited including 1) temporal smoothness~\cite{akhter2011trajectory, gotardo2011computing, kumar2016multi, kumar2018scalable},
2) fixed articulation~\cite{ramakrishna2012reconstructing} and more commonly used
3) shape priors. The first statistical shape prior---non-rigid objects can be modeled by a local
subspace in low rank---is first proposed by Bregler~\etal~\cite{bregler2000recovering} 
and later developed by Dai~\etal~\cite{dai2014simple}. Following this direction, increasing
works are reported to model more complex objects while still maintaining a well-conditioned system.
Among them, representatives are union-of-subspaces~\cite{zhu2014complex, agudo2018image}, and 
block-sparsity~\cite{kong2016prior, kong2016sfc}. Of particular interest to this paper is the most 
recent work~\cite{ck19} that introduces deep neural network to accurately solving large scale 
NRSfM problem. Even though great success, majority NRSfM algorithms rely heavily on 2D annotation-based 
priors. However, as pointed in the introduction, much broader information are embedded under image itself, 
under pixel values. In this paper, we impose a novel image prior such that NRSfM is no longer trapped at
2D coordinates of landmarks but also learn from origin images.

\paragraph{Weakly supervised 3D pose learning}

Most 3D pose estimation methods~\cite{integral, sup1, sup2, sup3, sup4, sup5, sup6, sup7, sup8, sup9} are fully supervised. One bottleneck for the supervised methods is that data coming from multi-view motion capture systems~\cite{joo2015panoptic,h36m} includes limited number of human subject, and has simple backgrounds. This would affect the generalization ability of a trained model. Weakly supervised methods aim to alleviate this problem by limiting the requirement for labeled data. They can be loosely categorized as: using synthetic datasets~\cite{w_chen2016synthesizing,w_varol2017learning} to increase the training set size. These methods face the problem of generalizing to new motions and environments that are different from the simulated data; On the other hand, given the existing large-scale image datasets with 2D annotation, Zhou~\etal~\cite{sup3} train their model with 2D labeled images together with motion capture data. To further reduce dependency on paired 3D annotation, 3D interpreter network~\cite{w_3dinterp}, multi-modal model~\cite{Tome_2017_CVPR} and generative adversarial networks~\cite{w_aign,w_repnet} are trained on external 3D data; multi-view footage is also used to enforce geometric constraints~\cite{tung2017self, w_Rhodin_cvpr}; However, these methods still require a large enough 3D training set to properly initialize and constraint their learning process. 

Recently, Rhodin \etal~\cite{w_Rhodin_eccv} propose a method based on geometric-aware representation learning, which requires only a small amount of annotation. Its performance however is limited, which restricts its practical usage. A concurrent work of Drover \etal~\cite{drover2018can} propose to use adversarial framework to impose a prior on the 3D structure, learned solely from 2D projections. Yet they still utilize the ground-truth 3D poses to generate a large number of synthetic 2D poses for training, which augments the original 1.5M 2D poses in Human3.6M by almost 10 times. 

% Compared to those, our method only requires cheap 2D annotation, and is able to provide better accuracy. 

% adversarial framework to impose a prior on the
% 3D structure, learned solely from their random 2D projections. Given
% a set of 2D pose landmarks, the generator network hypothesizes their
% depths to obtain a 3D skeleton. We propose a novel Random Projection
% layer, which randomly projects the generated 3D skeleton and sends the
% resulting 2D pose to the discriminator. The discriminator improves by
% discriminating between the generated poses and pose samples from a
% real distribution of 2D poses

% Drover \etal(ECCVW18) still utilizes the ground-truth 3D poses to generate a large number of synthetic 2D poses for training, which augments the original 1.5M 2D poses in Human3.6M by almost 10 times. In spite of this, both works show similar accuracy (65.5 vs 64.6 mm PA-MPJPE) if directly comparable (our prediction from RGB images vs theirs from detected 2d keypoints), but take a drastically different approach -- ours is developed from NRSfM while theirs uses GANs. Moreover, we demonstrated the benefits of associating with texture, which Drover \etal did not. For future work, it would be interesting to see the combination of both approaches to further close the gap between supervised and weakly-supervised methods.
\section{Non-rigid Structure from Motion}
\label{sec:nrsfm}
Under weak perspective camera assumption, 2D projection $\W\in\mathbb{R}^{P\times2}$ is the product of 3D shape $\S\in\mathbb{R}^{P\times3}$ and camera matrix $\M\in\mathbb{R}^{3\times2}$:
\begin{equation}
    \W = \mathbf{S}\mathbf{M},~~
    \W = \begin{bmatrix}\vdots & \vdots \\
    u_p & v_p\\
    \vdots & \vdots\end{bmatrix}, ~~
    \mathbf{S} = \begin{bmatrix} 
    \vdots & \vdots & \vdots\\
    x_p & y_p & z_p\\
    \vdots & \vdots & \vdots \end{bmatrix},
\end{equation}
where $(u_p, v_p)$ and $(x_p, y_p, z_p)$ are the image and world coordinate of $p$-th point, and $\M$  is required to be orthonormal. The goal of NRSfM is to recover 3D shape $\mathbf{S}$ and camera matrix $\M$ given the observed 2D projections $\W$. This is an inherent ill-posed problem. Finding a unique solution requires sufficient regularization and prior knowledge. 

One type of NRSfM methods approach the problem through dictionary learning. Denote $\mathbf{s}\in \mathbb{R}^{3P}$ is the vectorization of $\mathbf{S}$, it satisfies: $\mathbf{s} = \D\bvarphi$, where $\D\in\mathbb{R}^{3P\times K}$ is a dictionary with $K$ bases; and $\bvarphi\in\mathbb{R}^K$ is a code vector. Given multiple observation of 2D projections $\W^\supi$ from an articulated object deforming over time, or different objects of the same category, these methods can be loosely interpreted as minimizing the following objective:

\begin{equation}
    \min_{\D, \{\bvarphi^\supi\}, \{\M^\supi\}} \sum_i \|[\D\bvarphi^\supi]_{P\times3}\M^\supi-\W^\supi\|+h(\bvarphi^\supi)
    \label{eq:dict_nrsfm}
\end{equation}
where operator $[~]_{P\times3}$ is defined as reshaping the vecorized 3D shape into matrix form with dimension $P\times 3$; $h(\bvarphi)$ is a regularizer introduced to improve uniqueness of solution, e.g. low rank ~\cite{dai2014simple}, sparsity~\cite{kong2016prior}, \etc.

Our knowledge distilling method (see Section~\ref{sec:distill}) is designed for this general type of NRSfM method, and in principal, it is agnostic to the type of regularizor they use, as long as the dictionary is overcomplete.

\paragraph{Deep NRSfM}
Kong and Lucey\cite{ck19} propose a prior assumption that 3D shapes are compressible via multi-layer sparse coding:
\begin{equation}
\begin{aligned}
\mathbf{s} = \D_1\bvarphi_1,~~~&\|\bvarphi_1\|_1 \leq \lambda_1, & \bvarphi_1\geq0,\\
\bvarphi_1 = \D_2\bvarphi_2,~~~&\|\bvarphi_2\|_1 \leq \lambda_2, & \bvarphi_2\geq0,\\
\vdots, ~~~& \vdots & \\
\bvarphi_{n-1} = \D_n\bvarphi_n,~~~&\|\bvarphi_n\|_1 \leq \lambda_n, & \bvarphi_n\geq0,\\
\end{aligned}
\end{equation}
where $\D_i$ are hierarchical dictionaries, and code vectors $\bvarphi_i\in\mathbb{R}^{K_i}$ are constrained to be sparse and non-negative. Compared to single level sparse coding, codes in multi-layer sparse coding not only minimizes the reconstruction error at their individual levels, but is also regularized by the codes from other levels. This helps to impose more constraints on code recovery while maintaining similar shape expressibility versus single level sparse coding with the same dictionary size.

To recover sparse codes, one of the classical method to use is Iterative Shrinkage and Thresholding Algorithm (ISTA)~\cite{ista1,ista2,ista3}. Papyan \etal~\cite{papyan2017convolutional} find that feed-forward neural netorks can be interpreted as approximating one iteration of inferencing sparse codes by ISTA, and the dictionaries $\D_1,\D_2,\dots,\D_n$ serves as the neural network weights. Based on this insight, Chen \etal derived a novel neural network architecture which approximates the solution of sparse codes $\bvarphi_1$ and camera matrix $\M$. In this paper, we made significant modification to their original architecture, which we find important to get good result in experiment.  Limited by space, we put description about our version of camera matrix estimation network $q_\M(\W): \mathbb{R}^{P\times2}\mapsto \mathbb{R}^{3\times2} $, and sparse code estimation network $q_{\bvarphi}(\W, \M): \mathbb{R}^{P\times2}\times\mathbb{R}^{3\times2} \mapsto \mathbb{R}^{K_1}$ in the supplementary material. % Here, we emphasize a few modification we made on Chen's approach which we find important to get good result in experiment: 1) remove ReLU from pose estimation network; 2) instead of estimating code in the same forward pass with camera matrix, we use estimated camera matrix to first rotate the dictionary of first level, then  %section~\ref{sec:cam_est}, \ref{sec:code_est}. 

With the feed-forward code/camera estimation networks parameterized by the dictionaries, we can now learn the dictionaries through minimizing reprojection error of all samples in the dataset. Denote ${\tilde{\bvarphi}}_1^\supi$, $\tilde{\M}^\supi$ to be the output of networks $q_{\bvarphi}$, $q_{\M}$ given $i$th 2D projection $\W^\supi$, the loss function is:
\begin{equation}
    \min_{\D_1, \D_2, \dots, \D_n}\sum_i \|[\D_1 {\tilde{\bvarphi}}_1^\supi ]_{P\times3}\tilde{\M}^\supi - \W^\supi\|_2 + \lambda \|{\tilde{\bvarphi}}_1\|_1.
\end{equation}
In this loss function, in addition to reprojection error, we add sparisity penalty using a small weighting, which we find helpful to improve results.

% \cy{add a quick summary of our modification vs Chen \etal.}

% Our pose/code estimation networks are parameterized by the dictionaries $\D_1,\D_2,\dots,\D_n$

% \begin{equation}
% \begin{aligned}
%      & \min_{\mathbf{D}, \M_i, \bvarphi_i, \btheta} \sum_i \|\D^\sharp ( \bvarphi \otimes \M_i) - \W_i\|  + h(\bvarphi_i)\\
%     % \text{s.t.~~~} & \bvarphi_i \geq 0 \\
%     % & f_z(\mathbf{I}_i; \btheta) = \D^\sharp (\bvarphi \otimes \M_i^\perp)
% \end{aligned}
% \label{eq:nrsfm}
% \end{equation}

\section{Distilling Knowledge from NRSfM}
\label{sec:distill}

\begin{figure*}[t]
    \centering
    \includegraphics[width=\linewidth]{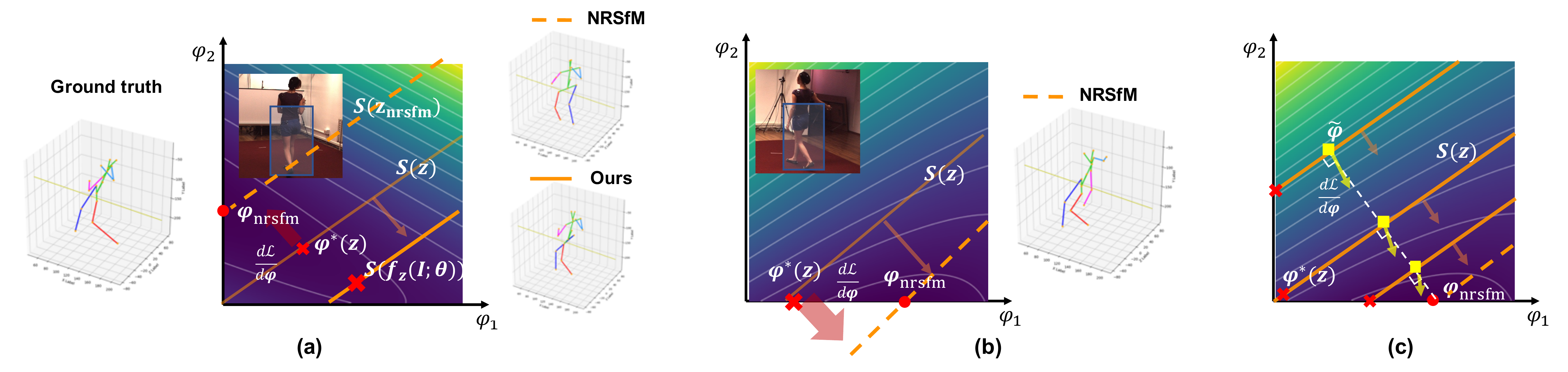}
    \caption{Illustration of the proposed knowledge distilling algorithm. \textbf{(a)} For illustration purpose, we assume the code $\bvarphi$ is 2-dimensional. We plot the cost function (Eq.~\ref{eq:code_loss}) as a 2D heatmap. The NRSfM solution $\bvarphi_\text{nrsfm}$ is approximately the minima of this heat map (represented as red dot). Given a depth hypothesis $\bz$, all the codes satisfies $\bz$ forms a subspace $\mathcal{S}(\bz)$, which is shown as the orange line. The quality of a depth hypothesis is evaluated by the best point on its subspace, denoted as $\bvarphi^*(\bz)$ (red cross). Given different depth hypothesis is equivalent to parallel translate the line. Suppose $\bz$ is free to have any value, then minimizing our loss function (Eq.~\ref{eq:exact_loss}) would push the line to cross $\bvarphi_\text{nrsfm}$(see the dashed orange line). This gives the same wrong depth reconstruction as the NRSfM method. (\textbf{b}) Suppose we get another image of similar pose but with less 2D projection ambiguity. In this case, NRSfM gives correct shape recovery. Since texture features are similar for both images, the pose estimation network is implicitly constrained to make similar depth predictions. Then minimizing our loss for both images would lead to a better solution for image 1 (shown as solid orange line), because gradients are larger from the 2nd image due to the fact that it has less ambiguity. \textbf{(c)} We approximate the loss by evaluating at the projection of $\bvarphi_\text{nrsfm}$ on the subspace (yellow square). This approximation is a convex upper bound for the original loss. It would still reflect the degree of projection ambiguity, and push the subspace (lines) torwards $\bvarphi_\text{nrsfm}$. 
    }
    \label{fig:intro_illu}
\end{figure*}

\noindent\textbf{Problem setup}: Given an image dataset paired with annotated 2D locations of landmarks on target objects: $\{(\mathbf{I}^\supi, \W^\supi)\}$, we want to train a 3D pose estimation network able to predict 3D landmark positions from image input. The main difficulty of this task is how to learn to predict depth of landmarks without any depth supervision. Our cue is from dictionary learning-based NRSfM method (Deep-NRSfM in our experiment), which gives us a 3D shape dictionary $\D$, and recovered camera matrices $\M^\supi$ and codes $\bvarphi_\text{nrsfm}^\supi$.

With the dictionary, camera matrices and codes from NRSfM, depth in the image coordinate can be computed by simply rotating the 3D shape reconstruction $\D\bvarphi_\text{nrsfm}^\supi$. Given this, a simple baseline for this task would be:  we use the depth reconstruction as labels to train the 3D pose estimation network. However, as shown in Fig.~\ref{fig:teaser}, we find that NRSfM tends to make wrong estimation due to strong ambiguity in 2D projections. Using those as hard target for regression would bottleneck the accuracy of learned pose estimation network. We propose a better approach - we want to establish a direct relation between depth prediction and the cost function (Eq.~\ref{eq:dict_nrsfm}) we used in NRSfM, which is the better metric to evaluate the quality of predicted 3D shapes. In this way, we can avoid confusing our student network with wrong labels, and allow them to implicitly associate image features to disambiguate difficult poses for NRSfM. This intuition is inline with other geometric self-supervised learning, e.g. self-supervised depth estimation~\cite{zhou2017unsupervised, godard2017unsupervised, wang2018learning}, in which photometric loss is used to train a depth estimation network.

\noindent\textbf{Outline}: The core problem is how to design a loss function which properly evaluates the quality of a depth hypothesis produced by the pose estimator. To derive our loss function, We first show that a depth hypothesis associates with a subspace of codes (see Section~\ref{sec:wsup_sec1}). We then advocate that the loss should be the minimum cost value of codes in the subspace (see Section~\ref{sec:wsup_sec2}). Finally, we derive a convex upper bound for the loss, which is computationally trackable for SGD training (see Section ~\ref{sec:wsup_sec3}). A 2D illustration is given in Fig.~\ref{fig:intro_illu} to help decipher the text.

\subsection{Depth hypothesis defines a subspace of codes}
\label{sec:wsup_sec1}
From NRSfM, we get the dictionary $\D$, and per example camera matrix $\M^\supi$. We find that the camera matrices from our modified Deep-NRSfM are accurate, thus we treat them as oracle and fixed in our learning algorithm. With this, we can simplify our notation by absorbing camera matrix into dictionary through rotation.  Rotation matrix $\mathbf{R}^\supi \in \mathbb{R}^{3\times3}$ is formed from camera matrix by:
\begin{equation}
    \R^{(i)} = [\mathbf{m}_1^{(i)}, \mathbf{m}_2^{(i)}, \mathbf{m}_1^{(i)} \times \mathbf{m}_2^{(i)}],
\end{equation}
where $\mathbf{m}_1^\supi$, $\mathbf{m}_2^\supi$ are columns of camera matrix $\M^\supi$. Then the dictionary is rotated by multiplying every 3D coordinates inside $\D$ with $\mathbf{R}^\supi$:
\begin{equation}
    \B^{(i)} = \begin{bmatrix}
    [\mathbf{d}_x^1, \mathbf{d}_y^1, \mathbf{d}_z^1]\R^{(i)} & \dots & [\mathbf{d}_x^P, \mathbf{d}_y^P, \mathbf{d}_z^P]\R^{(i)}
    \end{bmatrix}^T
\end{equation}

We further split $\B^{(i)}$ into two matrices -- one matrix takes all the $x$, $y$ coordinate elements of $\B^{(i)}$, while the other takes all the rest $z$ coordinate elements.
\begin{equation}
\begin{aligned}
    \mathbf{B}_{xy}^{(i)} = & \begin{bmatrix} 
    {\mathbf{b}_x^{1}}^{(i)} &
    {\mathbf{b}_y^{1}}^{(i)} & 
    \dots &
    {\mathbf{b}_x^{P}}^{(i)} &
    {\mathbf{b}_y^{P}}^{(i)} & 
    \end{bmatrix}^T,    \\
    \mathbf{B}_z^{(i)} = & \begin{bmatrix} 
    {\mathbf{b}_z^{1}}^{(i)} &
    \dots &
    {\mathbf{b}_z^{P}}^{(i)} &
    \end{bmatrix}^T,    \\   
\end{aligned}
\end{equation}
With this, $\mathbf{B}_{xy}^\supi \bvarphi^\supi$ computes 2D projection of shape reconstructed by code $\bvarphi^\supi$; and $\mathbf{B}_{z}^\supi \bvarphi^\supi$ is reconstructed depth in the image coordinate.

% \begin{equation}
%     \min_{\bvarphi^{(i)}}\|\mathbf{A}^{(i)}\bvarphi^{(i)} - \mathbf{w^{(i)}}\|_2 + \lambda \|\bvarphi^{(i)}\|_1
% \end{equation}
For a depth hypothesis $\mathbf{z'} = f_z(\mathbf{I}^{(i)}; \btheta)$ produced by the pose estimation network, codes giving depth reconstruction equal to $\mathbf{z}'$ forms a subspace:
\begin{equation}
    \mathcal{S}^\supi(\mathbf{z'}) = \{\bvarphi: \mathbf{B}_z^{(i)}\bvarphi = \mathbf{z}'\}.
\end{equation}
The subspace is not empty assuming that dictionary is overcomplete. In Fig.~\ref{fig:intro_illu}, the subspaces are visualized as orange lines in 2D.

\subsection{Loss = minimum cost on subspace}
\label{sec:wsup_sec2}
The quality of a depth hypothesis $\mathbf{z}'$ could be represented by the best code inside its subspace. As in NRSfM, the quality of a code is measured by the cost function = reprojection error + some regularizer, i.e.:
\begin{equation}
    \mathcal{C}^\supi(\bvarphi) = \|\mathbf{B}_{xy}^{(i)}\bvarphi - \mathbf{w}^{(i)}\| +  h(\bvarphi),
    \label{eq:code_loss}
\end{equation}
where $\mathbf{w}^{(i)}$ is the vectorization of $\W^\supi$. To keep formulation general, we don't specify the type of norm and regularizer here. Thereby we have the following definition of quality function for $\mathbf{z}'$, which we use as the loss function for knowledge distilling:
% \begin{equation}
%     \cL(\mathbf{z}'^{(i)}) = \min_{\bvarphi\in \mathcal{S}(\mathbf{z}'^{(i)})} \|\mathbf{B}_{xy}^{(i)}\bvarphi - \mathbf{w}^{(i)}\| +  h(\bvarphi)
%     \label{eq:exact_loss}
% \end{equation}
\begin{equation}
    \cL^\supi(\mathbf{z}') = \min_{\bvarphi\in \mathcal{S}^\supi(\mathbf{z}')} \mathcal{C}^\supi(\bvarphi).
    \label{eq:exact_loss}
\end{equation}
This computes the minimum cost value of codes inside the subspace defined by the depth hypothesis $\mathbf{z}'$. 

To evaluate this loss function, we need to first solve for the minima $\bvarphi^*$ of the constrained convex optimization problem in Eq.~\ref{eq:exact_loss} (red cross in Fig.~\ref{fig:intro_illu}). Suppose we can express $\bvarphi^*$ as a differentiable function of $\mathbf{z}'$, i.e. $\bvarphi^*=q^\supi(\mathbf{z}')$,  Eq.~\ref{eq:exact_loss} becomes:
\begin{equation}
     \cL^\supi(\mathbf{z}') = \|\mathbf{B}_{xy}^{(i)}q^\supi(\mathbf{z}') - \bw^\supi \| + h(q^\supi(\mathbf{z}')).
     \label{eq:exact_loss2}
\end{equation}
This loss is explicitly a function of $\mathbf{z}'$, and thus allows the gradients to be propagated to the pose estimation network.

As a side note, suppose the pose network has unlimited capacity, in other words, able to overfit any depth values, then the end result of minimizing this loss function would be a network predicting the same depth as the NRSfM algorithm (illustrated in Fig.~\ref{fig:intro_illu}(a)). We argue that this would not be the case in practice, since convolution networks constrained by their structure, is equivalent to have a deep image prior~\cite{ulyanov2018deep} imposed on their output. This image prior provides extra constraint to disambiguate confusing 2D projections, thus is the key source for our improvement over the NRSfM teacher.

\subsection{Convex upper bound of Eq.~\ref{eq:exact_loss2}}
\label{sec:wsup_sec3}
Using Eq.~\ref{eq:exact_loss2} requires to form the (sub)differentiable function $q^\supi(\mathbf{z}')$ which produces the solution to the constrained optimization problem in Eq.~\ref{eq:exact_loss}. However, solving this constrained optimization problem requires iterative numerical method due to the existence of regularizer. As a result, it's computationally intractable to solve it exactly per SGD iteration during training. Therefore we derive an approximate solution as follow:

Suppose $\bvarphi_\text{nrsfm}^{(i)}$ is the solution we get from NRSfM, and it approximates the minima of the optimization problem in Eq.~\ref{eq:exact_loss} without the subspace constraint, then an approximate solution for the constrained problem could be the projection of $\bvarphi_\text{nrsfm}^{(i)}$ onto the subspace $\mathcal{S}^\supi(\mathbf{z}')$:
\begin{equation}
    \tilde{\bvarphi}^{(i)}(\mathbf{z}') = \arg\min_{\bvarphi\in \mathcal{S}^\supi(\mathbf{z}')} \frac{1}{2}\|\bvarphi - \bvarphi_\text{nrsfm}^{(i)}\|_2^2
    \label{eq:proj_loss}
\end{equation}
%B^T(BB^T)^-1
The closed form solution to Eq.~\ref{eq:proj_loss} is:
\begin{equation}
\tilde{\bvarphi}^{(i)}(\mathbf{z}') = \bvarphi_\text{nrsfm}^\supi + (\B_z^\supi)^\dagger (\mathbf{z}' -  \B_z^\supi \bvarphi_\text{nrsfm}^\supi ), 
\label{eq:proj}
\end{equation}
where $ (\B_z^\supi)^\dagger = {\B_z^\supi}^T(\B_z^\supi {\B_z^\supi}^T)^{-1}$ is the right inverse of $\B_z^\supi$.
Eq.~\ref{eq:proj} is implemented as a differentiable operator thanks to modern deep learning library.

Substitute the exact solution $q^\supi(\mathbf{z}')$ in Eq.~\ref{eq:exact_loss2} by the approximate solution $\tilde{\bvarphi}^{(i)}(\mathbf{z}')$ gives a convex upper bound of ~Eq.~\ref{eq:exact_loss2}: 
\begin{equation}
    \tilde{\cL}^\supi(\bz') = \|\B_{xy}^\supi \tilde{\bvarphi}^\supi(\bz') - \bw^\supi \| +  h(\tilde{\bvarphi}^\supi(\bz'))
    \label{eq:loss_ub}
\end{equation}
In our experiment, we find that using this convex upper bound as training loss, is sufficient to give lower error on the training set compared to our already strong NRSfM baseline.

\subsection{Learning the 3D pose estimator}
We use the state-of-the-art integral regression network~\cite{integral} as our student pose estimator. The network directly predicts 3D coordinates of landmarks in the image coordinate. During training, the $(x,y)$ coordinate is directly supervised by 2D landmark annotations; while $z$ coordinate is supervised by our knowledge distilling loss (Eq.~\ref{eq:loss_ub}). The proposed learning objective is:
\begin{equation}
    \min_{\btheta} \sum_i \|f_{xy}(\mathbf{I}^\supi; \btheta) - \bw^\supi\|_1 + \tilde{\cL}^\supi(f_z(\mathbf{I}^\supi; \btheta)),
\end{equation}
where $f_{xy}$,$f_z$ denote the output of the network at $(x,y)$ and $z$ coordinates; and $\btheta$ refers to the network weights. For the knowledge distilling loss $\tilde{\cL}$, we use $L_2$ norm for the reprojection error, and $L_1$ norm for the regularizer in our experiment. The regularizer is weighted by an empirically found coefficient, which is 0.3 in our experiment.

\section{Experiment}

\begin{table}[]
    \centering
    \begin{small}
    \begin{tabular}{l|c|c|c}
    \hline
         &  P-MPJPE & MPJPE & \shortstack{depth \\ error} \\
         \hline
    Ranklet~\cite{del2007non} & 281.1 & - & -\\
    Sparse~\cite{kong2016prior} & 217.4 & - & - \\
    SPM(2k)~\cite{dai2014simple} & 209.5 & - & -\\
    SFC~\cite{kong2016sfc} & 167.1 & 218.0 & 135.6\\
    KSTA(5k)~\cite{gotardo2011kernel} & 123.6 & - & -\\
    RIKS(5k)~\cite{hamsici2012learning} & 103.9 & - & -\\
    Consensus~\cite{lee2016consensus} & 79.6 & 120.1 & 111.5\\
    \hline
    Deep-NRSfM$^*$~\cite{ck19} & 73.2 & 101.6 & 76.5 \\
    Weaksup-bs & 61.2 & 86.2 & 75.3 \\
    Ours & 56.4 & 80.9 & 71.2 \\
    \hline
    \end{tabular}
    \end{small}
    \caption{Compare with NRSfM methods on the training set of H3.6M ECCV18 challenge dataset. KSTA, RIKS are evaluated on a subset of 5k images, and SPM is evaluated on 2k images. $*$ Our implementation of Deep-NRSfM has significant difference compared to the original paper.}
    \label{tab:nrsfm}
\end{table}

\begin{table}[h]
\footnotesize
    \centering
    \begin{tabular}{l|c|c|c|c|c}
    \hline
        & 2D & 3D & MV & P-MPJPE & MPJPE  \\
         \hline
    \hline
    Sun \etal~\cite{integral} & - & - & - & - & 86.4 \\
    \hline
    Rhodin \etal~\cite{w_Rhodin_eccv} & & \checkmark & \checkmark & 98.2 & 131.7 \\
    Tung \etal~\cite{tung2017self}& \checkmark & \checkmark & \checkmark & 98.4 & - \\
    3Dinterp.~\cite{w_3dinterp} & \checkmark & \checkmark & & 98.4 & - \\
    AIGN~\cite{w_aign} & \checkmark & \checkmark &  &  97.2 & -  \\
    Tome \etal~\cite{Tome_2017_CVPR} & \checkmark & \checkmark & & - & 88.4  \\
    Drover \etal~\cite{drover2018can} & \checkmark & \checkmark & & 64.6 & - \\
    % RepNet & \checkmark & \checkmark & & 65.1 & 89.9 & - \\
    \hline
    Weaksup-bs & \checkmark & & &  67.3 & 95.0  \\
    Ours & \checkmark & & &  62.8 & 86.4  \\
    + MPII & \checkmark & & & \textbf{57.5} & \textbf{83.0} \\
    % Ours & \checkmark & & &  70.5 & 96.2 & 73.1 \\
    % + MPII & \checkmark & & & 65.5 & 91.1 & 71.8 \\
    \hline
    \end{tabular}
    \caption{Compare with weakly supervised methods on H3.6M validation set. Supervision source used by each method is marked: `2D' refers to 2D landmark annotation; `3D' represents any training source with 3D annotation, including synthetic 3D dataset, external human 3D model, etc.; `MV' is the abbreviation for multi-view.}
    \label{tab:sup}
\end{table}

\begin{figure*}
    \centering
    \includegraphics[width=\linewidth]{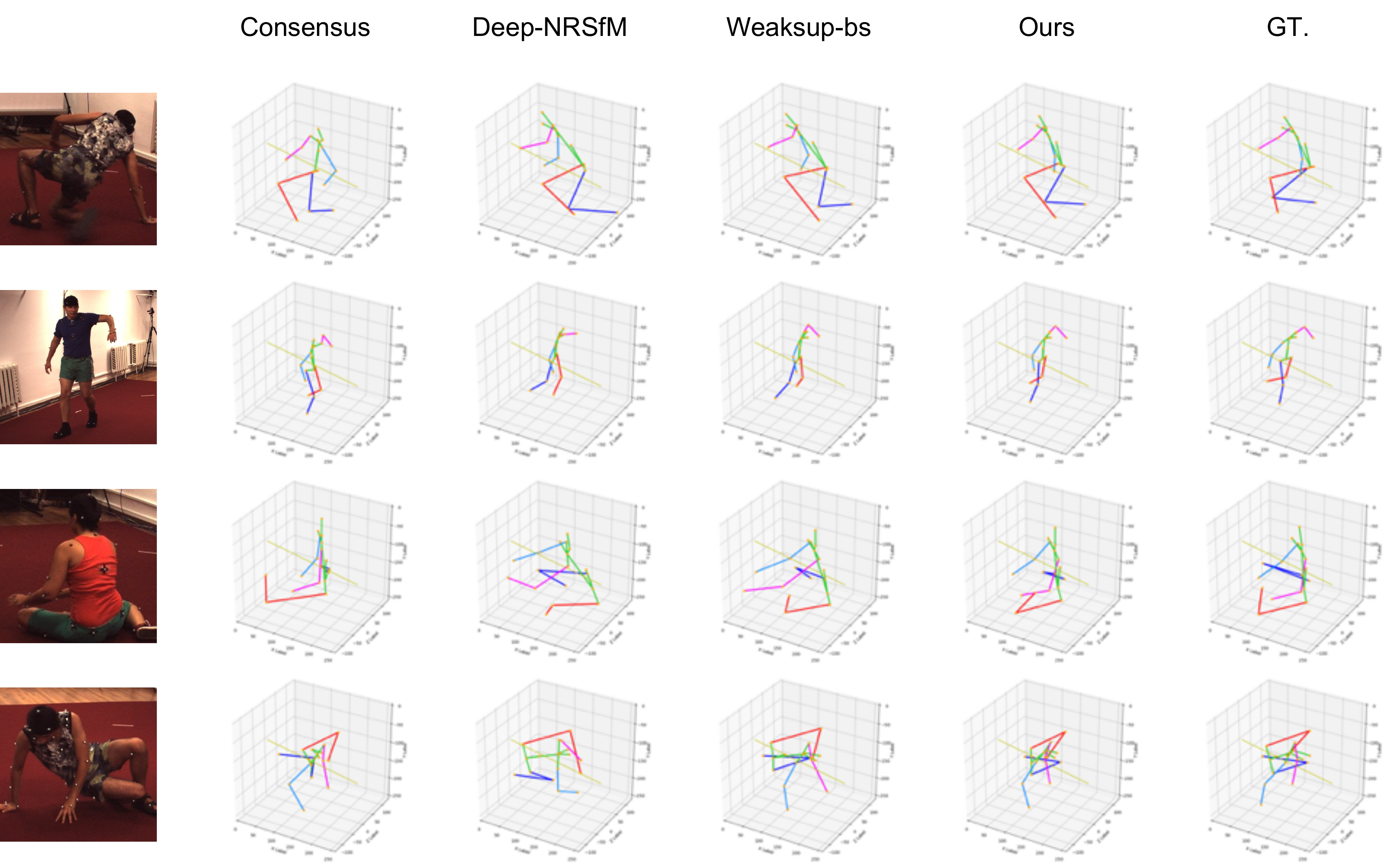}
    \caption{Visual comparison of NRSfM methods versus methods which include image as extra constraint (i.e. our weakly supervised baseline and our knowledge distilling method) on the training set. Our method shows significant improvement over its teacher, i.e. deep-NRSfM. Skeletons are rendered from side view for better visualization of the difference in depth reconstruction. We use red and magenta to color left leg and arm, while blue and dodgerblue are used to color right leg and arm.} 
    \label{fig:nrsfm_comp}
\end{figure*}

\begin{table*}
\begin{scriptsize}
    \centering
    \begin{tabular}{l|*{15}{c}}
    \hline
         & Direct. & Disc. & Eat & Greet & Phone & Photo & Pose & Purch. & Sit & SitD & Smoke & Wait & Walk & WalkD & WalkP\\
         \hline
      3Dinterp.~\cite{w_3dinterp}  & 78.6 & 90.8 & 92.5 & 89.4 & 108.9 & 112.4 & 77.1 & 106.7 & 127.4 & 139.0 & 103.4 & 91.4 & 79.1 & - & -\\
      AIGN~\cite{w_aign} & 77.6 & 91.4 & 89.9 & 88.0 & 107.3 & 110.1 & 75.9 & 107.5 & 124.2 & 137.8 & 102.2 & 90.3 & 78.6 & - & -\\
      Drover \etal~\cite{drover2018can} &60.2 & 60.7 & 59.2 & 65.1 & 65.5 & 63.8 & 59.4 & 59.4 & 69.1 & 88.0 & 64.8 & 60.8 & 64.9 & 63.9 & 65.2\\
    %   RepNet & \textbf{53.0} & \underline{58.3} & \textbf{59.6} & 66.5 & 72.8 & 71.0 & \textbf{56.7} & \textbf{69.6} & \textbf{78.3} & 95.2 & \textbf{66.6} & \textbf{58.5} & \underline{63.2}\\
      \hline
      Weaksup-bs & 58.8 & 62.4 & 56.7 & 59.8 & 68.6 & 60.8 & 59.7 & 81.0 & 93.4 & 68.5 & 75.8 & 65.9 & 61.5 & 67.6 & 65.0\\
      Ours & 54.7 & 57.7 & 54.8 & 55.8 & 61.6 & 56.3 & 52.7 & 73.7 & 95.5 & 62.3 & 68.5 & 60.8 & 55.5 & 64.0 & 58.0\\
      +MPII & \textbf{50.3} & \textbf{48.9} & \textbf{52.7} & \textbf{53.9} & \textbf{59.9} & \textbf{50.7} & \textbf{48.3} & 70.9 & 82.6 & \textbf{58.0} & 65.3 & \textbf{54.7} & \textbf{50.8} & \textbf{57.7} & \textbf{55.6}\\
    %   Weaksup-bs & 62.0 & 68.7 & 65.3 & 65.4 & 74.7 &  78.8 & 63.2 & 68.9 & 87.6 & 146.1 & 72.7  & 71.2 & 64.8\\
    %   Ours & 59.6 & 66.2 & 60.2 & 62.4 & 69.1 & 75.5  & 59.1 & 71.0 & 80.3 & 113.4 & 69.8  & 66.3 & 63.6\\ 
    %   +MPII & \textbf{53.7} & \textbf{57.2} & \textbf{60.4} & \textbf{59.7} & \textbf{66.4} & \textbf{72.3} & \textbf{53.5} & \textbf{62.6} & \textbf{79.7} & \textbf{102.4} & \textbf{65.4} & \textbf{60.1} & \textbf{58.8}\\
      \hline
    \end{tabular}
    \end{scriptsize}
    \caption{Per action PA-MPJPE reported on H3.6M validation set. Our approach performs favorably compared to other weakly supervised methods.}
    \label{tab:per_act}
\end{table*}

\subsection{Implementation details}
\noindent\textbf{Data preprocessing:} We assume no knowledge of 3D label in both training and testing. We crop the image according to the 2D human bounding box, and then resize and pad such that it is 256x256 resolution. The 2D points are then represented by the patch coordinate. In evaluation, we follow the same procedure as in~\cite{integral}, which aligns the scale of the prediction by average bone length before computing the metrics. \\
\noindent\textbf{3D pose estimation network:} We select the integral regression network~\cite{integral} due to its state-of-the-art performance in human pose estimation. Throughout our experiment, we use ResNet50 as the backbone for the regression network, and the input image resolution is set as $256\times256$. Using deeper backbone network (e.g. ResNet152) and higher image resolution would improve result, as already shown in~\cite{integral}. We choose this cheaper setting for a fairer comparison with other weakly supervised methods which use ResNet50.\\

During training, we follow most of the settings in~\cite{integral}, i.e. the base learning rate is 1e-3, and it drops to 1e-5 when the loss on the validation set saturates. Limited by our computational resources, we use a smaller batch size of 32.

\noindent\textbf{Deep-NRSfM:} We use dictionaries with 6 levels. The size for the dictionaries from lower level to higher is: 256, 128, 64, 32, 16, 8. When learning the dictionaries, the sparsity weight ($\lambda$ in Eq.~\ref{eq:dict_nrsfm}) is selected through cross validation and set as 0.01. For more details of our modified version of Deep-NRSfM, we refer the reader to our supplementary material.

\subsection{Experiment setup}
\noindent\textbf{Dataset:} We validate our method on Human3.6M dataset (H3.6M)~\cite{h36m}, which is the major dataset used in current 3D human pose estimation research. Despite our experiment is focused on human pose estimation, we'd like to emphasize that the proposed method is a general algorithm. Unlike other weakly supervised methods which are deeply coupled with external 3D human model, our method doesn't require any target specific prior knowledge, thus should be applicable to other type of objects without restriction.

H3.6M includes sequences of 11 actors performing 15 type of actions captured from 4 camera locations. Footage of 7 out of 11 actors are released for training/validation. We follow the experiment convention conducted by prior papers: 5 subjects (S1, S5, S6, S7, S8) are used as training set, and 2 subjects (S9, S11) for testing. Although H3.6M dataset comes with 3D annotation, we use only 2D annotation during training, and 3D labels are kept for validation.

Strategies to sample frames from the training footage can have a direct impact on validation accuracy. For reproducibility, we use the subset (35k+ images) selected by H3.6M ECCV18 Challenge for training. We augment the training set through random image warping and perturbation as in ~\cite{integral}.

\noindent\textbf{Evaluation metric:}
We follow the two common evaluation protocols used in literature, and report both of them.
\begin{itemize}
    \item MPJPE: mean per joint positioning error measures the mean euclidean distance between the reconstructed and ground truth joints after shifting them to have the same root joint coordinate.
    \item PA-MPJPE: Align the reconstructed joints to the ground truth through rigid transformation before evaluating MPJPE. This metric is more often used in NRSfM to measure the correctness of the reconstructed shape.
\end{itemize}
In addition, we also report `depth error' which measures the mean difference along z-axis. This is the most important metric to validate our method, because the core problem of weakly supervised learning is how to recover depth without annotation.

\noindent\textbf{Weakly supervised learning baseline:}
As previously mentioned, a simple weakly supervised learning baseline is using the depth output from our Deep-NRSfM method as training labels. We use this baseline (refer as ``Weaksup-bs'') to validate the contribution of our novel knowledge distilling loss. To train the pose estimation network, we employ L1 regression loss which has been proven effective in~\cite{integral}.

\noindent\textbf{Weighting value for the $L_1$ regularizer:}
We study the effect of different weighting values for the $L_1$ regularizer in the propoesed knowledge distilling loss (Eq.~\ref{eq:loss_ub}). As shown in Table~\ref{tab:l1_weighting}, under a reasonable range (0.1-0.5) of the weights, our method consistently outperforms the baseline.
\begin{table}[h]
    \centering
    \begin{small}
    \begin{tabular}{c|c|c|c|c|c}
    \hline
         $L_1$weight &  0.01 & 0.1 & 0.3 & 0.5 & Weaksup-bs\\
          \hline
         depth error (mm) & 79.0 & 74.6 & \textbf{73.1} & 76.7 &  78.0\\
         PA-MPJPE (mm) & 73.0 & 73.6 & \textbf{70.5} & 71.0 & 75.8\\
    \hline
    \end{tabular}
    \end{small}
    \caption{Comparing different weighting values for the $L_1$ regularizer in Eq.~\ref{eq:loss_ub}. Numbers reported on the  validation  set of H3.6M ECCV18 challenge.}
    \label{tab:l1_weighting}
\end{table}

\paragraph{Using extra data from MPII:}
Prior works~\cite{sup3} has shown that including external 2D data such as MPII~\cite{mpii} as training source can improve generalization ability of the learned 3D pose estimator. Thus, we also report result of our method trained with H3.6M+MPII. Due to our current method does not handle missing joints, we apply our proposed knowledge distilling loss only to those MPII images with complete 2D skeleton annotation; for images with occluded/out-of-view joints, we only use 2D regression loss as in~\cite{integral}.

\begin{figure*}[t]
    \centering
    \includegraphics[width=.9\linewidth]{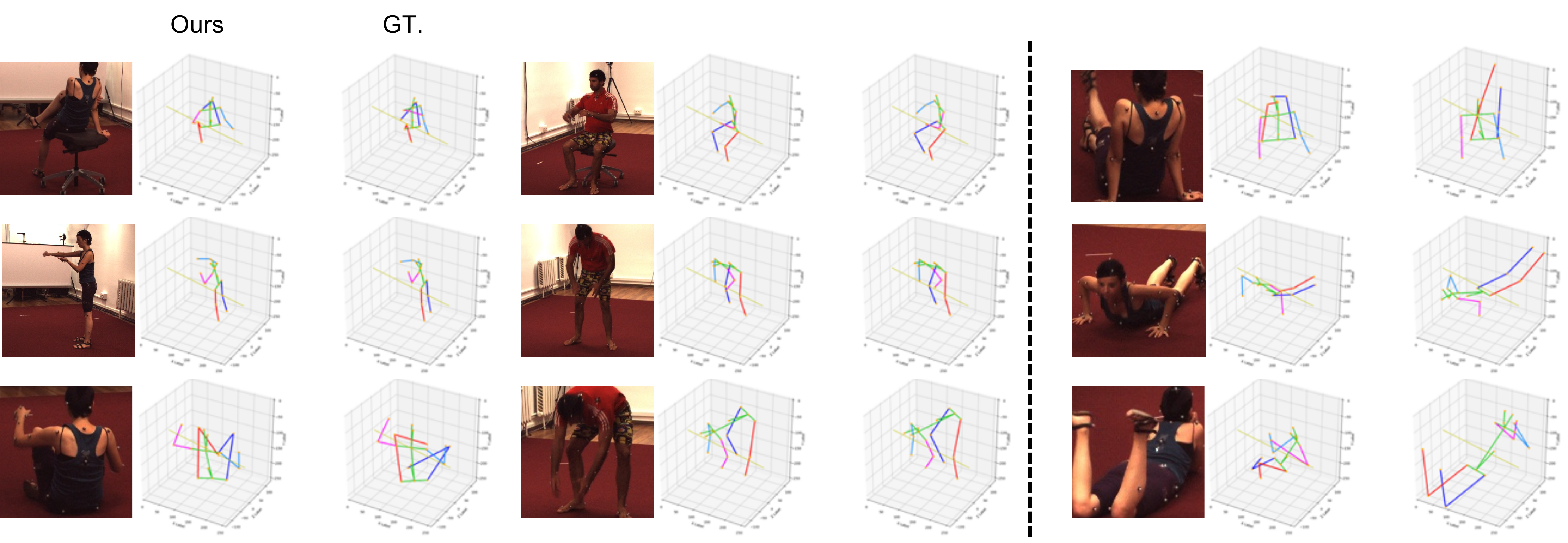}
    \caption{Qualitative results of ours on H3.6M validation set. The right part shows some of our failure cases. Our method may fail under severe occlusion and rare body poses.}
    \label{fig:sup_comp}
\end{figure*}

\subsection{Compare with NRSfM methods}

We compare with 7 state-of-the-art NRSfM methods on our training set (35k+ images from H3.6M ECCV18 Challenge). We find this dataset is challenging to the compared methods due to: 1) large variation in camera positions; 2) difficult poses such as sitting and prone occupy a significant portion of the dataset; 3) variation in scale is large, due to the fact that without the knowledge of 3D, we cannot normalize 2D projections by distance or calculating bone length. The best we can do is to normalize 2D points by the size of 2D bounding box. This leads to certain pose e.g. sitting appears larger compared to others after normalization; 4) some of the methods fails to cope with a large number of samples (e.g. $>$5k). For those methods, we report result on the largest subset they can handle. We also try to compare with the recently proposed MUS~\cite{agudo2018image}, but their implementation fails to handle H3.6M dataset with large number of frames.

Despite of these difficulties, our implementation of Deep-NRSfM  outperforms all of them. As shown in Table.~\ref{tab:nrsfm}, it reduces depth error by more than $33\%$ compared to the second best. This means that switching to other NRSfM method is bound to inferior result of training a 3D pose estimator.

More interestingly, although our weakly supervised learning baseline (Weaksup-bs) is trained to reconstruct the same depth value produced by deep NRSfM, it actually gets slightly lower depth error compared to its regression target. This indicates that the deep image prior is taking effect, but still restricted by the noisy labels from Deep-NRSfM.

Finally, the pose estimation network learned by our knowledge distilling loss reduces the depth error from Deep-NRSfM's 76.5mm to 71.2mm. As shown in Fig.~\ref{fig:nrsfm_comp} and ~\ref{fig:teaser}, this 5.3mm average difference includes a huge improvement in cases such as identifying if a leg is stretching towards or away from the camera.

\subsection{Compare with weakly supervised methods}

We compare with other weakly supervised 3D pose learning methods on the H3.6M validation set. In Table.~\ref{tab:sup}, we first list the performance of Integral regression network by Sun \etal~\cite{integral} as a supervised learning baseline. We copied its MPJPE (corresponding to ResNet50 with $256\times256$ input size and $I_1$ loss) from their paper. Since in our experiment, we're using exactly the same pose estimation network architecture, this serves as the upper bound of accuracy, which a weakly supervised learning method can achieve.

Next, we list results from 7 weakly supervised methods, and the type of their training source is marked. `2D' refers to 2D landmark annotation; `3D' represents any external 3D training source, including 3D human models, unpaired 3D skeleton dataset, synthetic dataset with 3D annotations, etc.; MV is the abbreviation for multi-view footage. We find that our method outperforms all the compared methods, while using the least amount of supervision. We also experiment with including MPII as extra training source, which leads to more error reduction. Fig.~\ref{fig:sup_comp} shows some qualitative results of our method on the validation set.
For per action error break down, we list PA-MPJPE of 13 different actions in Table~\ref{tab:per_act}.

\section{Conlusion}
In this paper, we presented a weakly supervised 3D pose learning algorithm requires zero 3D annotation. We proposed a novel loss to distill knowledge from a general type of NRSfM method based on dictionary learning. We also established a strong NRSfM baseline on a challenging dataset, beating all the state-of-the-arts. 
Despite its current sucess, the limitations of our method are: 1) we require weak perspective projection, thus objects with strong perspective change is not ideal for the proposed method; 2) we do not model missing labels yet, thus another iteration is needed to extend the method to datasets with lots of occluded/out-of-view objects. We leave these for future work.
{\small
\bibliographystyle{ieee_fullname.bst}
\bibliography{egbib}
}

\end{document}